\documentclass{article}

\usepackage[preprint]{neurips_2021} 

\usepackage[utf8]{inputenc} 
\usepackage[T1]{fontenc}    
\usepackage{hyperref}       
\usepackage{url}            
\usepackage{booktabs}       
\usepackage{amsfonts}       
\usepackage{nicefrac}       
\usepackage{microtype}      
\usepackage{xcolor}         
\usepackage{graphicx}
\usepackage{float}
\usepackage{multirow}
\usepackage{enumitem}
\usepackage[export]{adjustbox}
\graphicspath{ {./images/} }

\title{Using Synthetic Images To Uncover Population Biases In Facial Landmarks Detection}

\author{
  Ran Shadmi, \qquad Jonathan Laserson, \qquad Gil Elbaz\\
  Datagen Technologies Ltd\\
  Israel \\
  \texttt{\{ran.shadmi, jonathan, gil\}@datagen.tech} \\
}

\begin{document}

\maketitle

\begin{abstract}
  In order to analyze a trained model performance and identify its weak spots, one has to set aside a portion of the data for testing. The test set has to be large enough to detect statistically significant biases with respect to all the relevant sub-groups in the target population. This requirement may be difficult to satisfy, especially in data-hungry applications. We propose to overcome this difficulty by generating \textit{synthetic} test set. We use the face landmarks detection task to validate our proposal by showing that all the biases observed on real datasets are also seen on a carefully designed synthetic dataset. This shows that synthetic test sets can efficiently detect a model's weak spots and overcome limitations of real test set in terms of quantity and/or diversity.
\end{abstract}

\section{Introduction}

Human-centered computer vision is becoming ubiquitously adopted in every device and service that requires to identify and understand the humans users. These widely adopted algorithms may contain inherent biases that leak from their training data. Using a data-centric approach, we push towards quick iterations of testing our models to find these biases and then solve them. Synthetic data can be generated in large quantities and be engineered to contain any visual attribute, however rare it is in real life. Thus it enables faster and more focused iterations, especially when real data is scarce or difficult to collect.

\section{Related Work}

Synthetic data has shown to be useful in many machine-learning applications. It is most commonly used in the training process by augmenting the real training samples \cite{Kortylewski2018, DBLP:journals/corr/RichardsonSK16, DBLP:journals/corr/abs-1907-02499, DBLP:journals/corr/abs-1804-06516} when the available amount of real samples is small or when some important sub-groups are under-represented. In \cite{Buolamwini2018}, three commercial gender classification systems were shown to have weaker performance for darker-skinned females. In \cite{georgopoulos2020}, the KANFace richly annotated dataset was used to expose biases for age, gender and skin color in a variety of face analysis tasks.

Little work has been published on using synthetic data to augment or replace the \textit{test} dataset. Recently, simulated data has been used to test autonomous driving \cite{Meinke}. Tesla \cite{Tesla} uses synthetically generated data to create specific scenarios that occur rarely in real-life and test their models against it.

\section{Method}

In order to validate the hypothesis that \textit{synthetic} test set can uncover biases existing in \textit{real} test sets, we designed an experiment composed of the following steps:
\begin{enumerate}[leftmargin=*]
\item Use the widely-used DLIB package \cite{dlib09} to detect facial landmarks.
\item Use the CelebA \cite{liu2015faceattributes} and FFHQ-Aging \cite{DBLP:journals/corr/abs-2003-09764} real-faces datasets which have both \textbf{facial landmarks annotations} and various \textbf{appearance attributes} (such as gender, age and skin color).
\item Evaluate the trained model on the real datasets and measure its performance using the widely-used NME (Normalized Mean Error)\cite{Wang_2019} score.
\item Stratify the errors based on the appearance attributes to identify statistically significant biases, e.g. the model may have lower average error on women (vs. men) faces.
\item Create rich synthetic faces data using Datagen's faces platform \cite{datagen}, carefully controlled as to contain a significant amount of samples from each attribute.
\item Evaluate the trained model on the generated synthetic data using the same error measure and compare the biases to those observed on the real data.
\end{enumerate}

\begin{figure}
\centering
\begin{tabular}{cccc}
\centering
\includegraphics[width=.21\linewidth,valign=m]{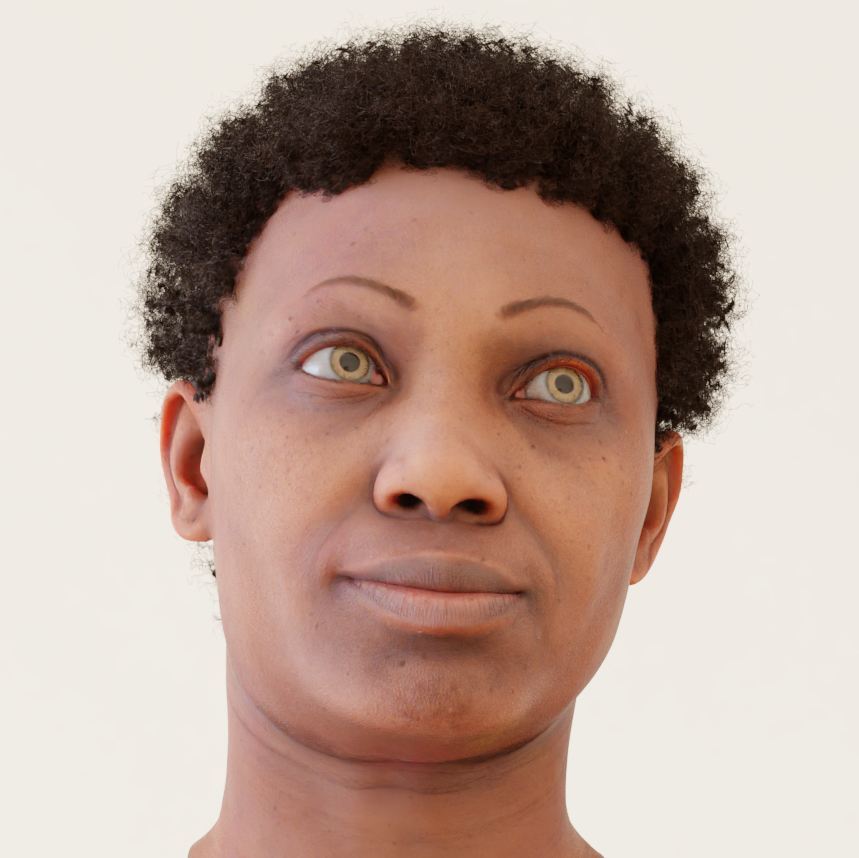} & \includegraphics[width=.21\linewidth,valign=m]{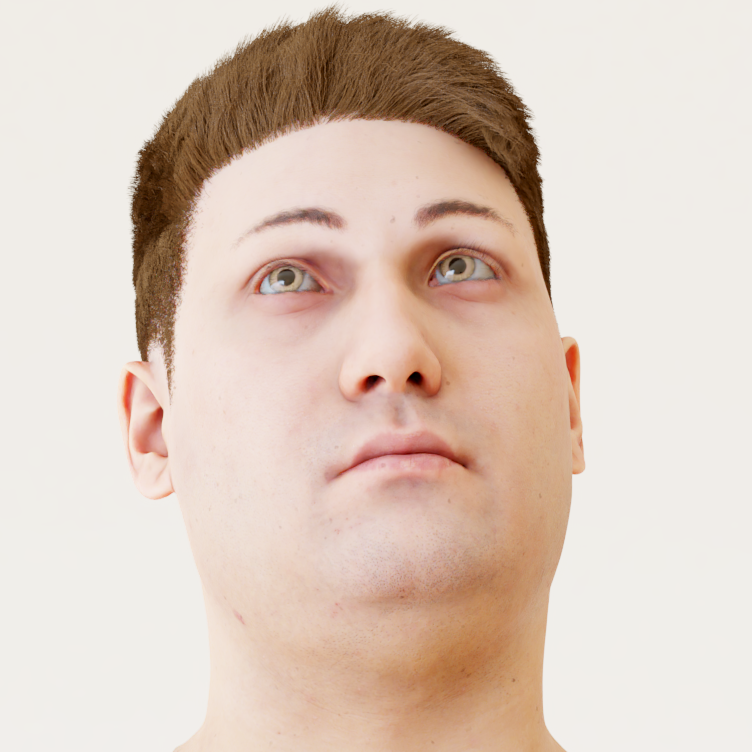} &
\includegraphics[width=.21\linewidth,valign=m]{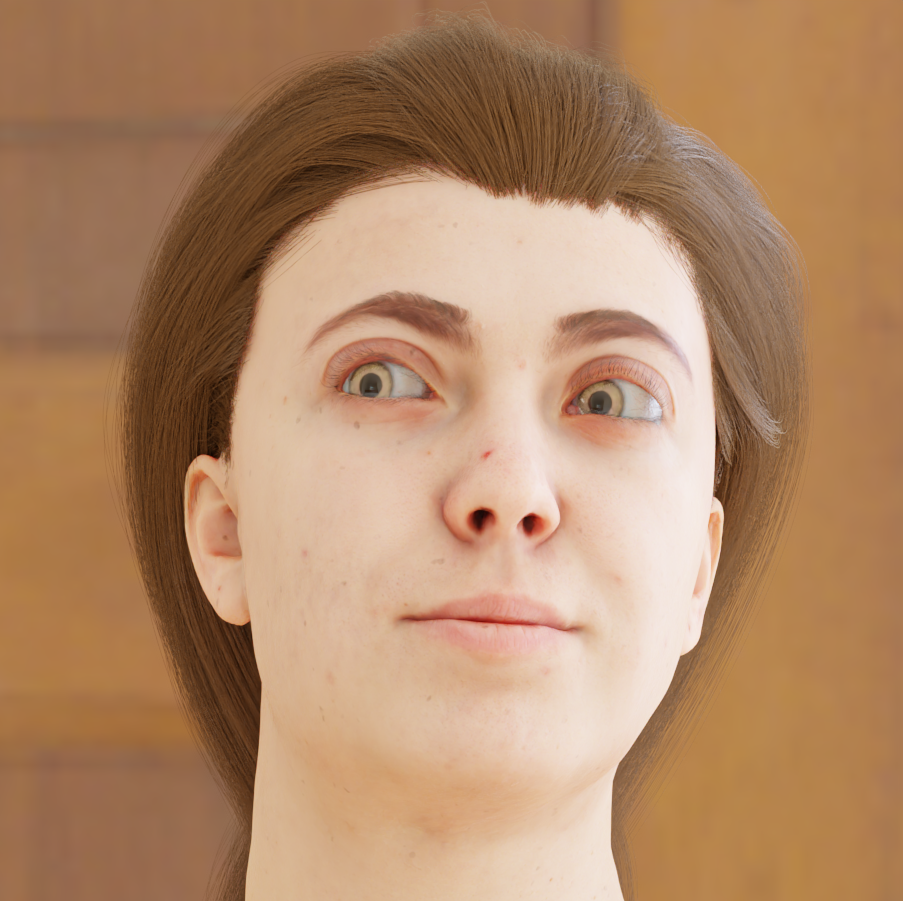} &
\includegraphics[width=.21\linewidth,valign=m]{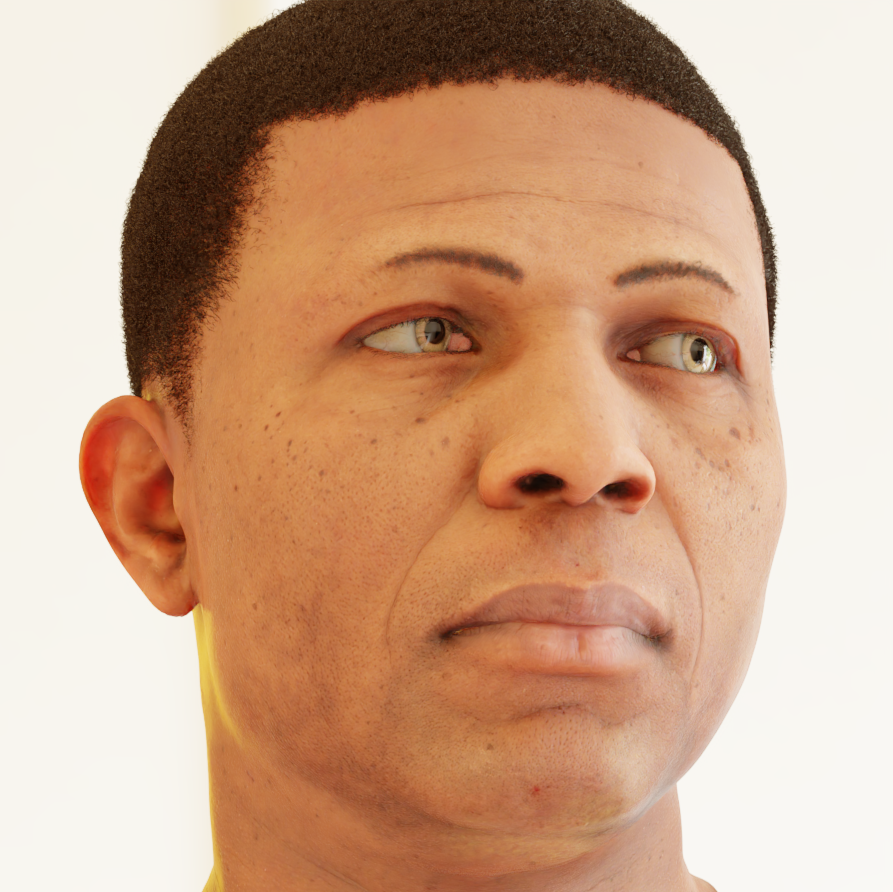}\\
\end{tabular}
\caption{Examples of face images from the Datagen synthetic dataset.}
\label{figure:face_images}
\end{figure}

\begin{table}
\centering
\caption{
    Facial landmarks detection results using DLIB for the CelebA / FFHQ-Aging / synthetic datasets and for various attributes. Each multi-row represents an appearance attribute (and its complement), each row within the multi-row represents the results on a specific data-set. All the results are statistically significant with very small \textit{P < .001}.
}
\label{table:results_table}
\begin{tabular}{c|c|cc|cc|r}
\toprule
attribute  & dataset    & \multicolumn{2}{|c|}{w/ attribute} & \multicolumn{2}{|c|}{w/o attribute} & \multicolumn{1}{c}{delta} \\
           &            & mean NME & \# samples              & mean NME & \# samples               & \\
\midrule
young      & CelebA     & 4.41\%  & 152,399                 & 4.81\%  &  43,955                  & -0.40\% \\
vs.        & FFHQ-Aging & 1.66\%  &  18,081                 & 1.87\%  &  30,402                  & -0.21\% \\
old        & synthetic  & 3.38\%  &   2,502                 & 4.16\%  &   3,018                  & -0.79\% \\
\hline
men        & CelebA     & 4.79\%  &  80,660                 & 4.29\%  & 115,694                  & 0.49\% \\
vs.        & FFHQ-Aging & 1.91\%  &  22,846                 & 1.69\%  &  25,637                  & 0.22\% \\
women      & synthetic  & 4.11\%  &   3,237                 & 3.37\%  &   2,283                  & 0.75\% \\
\hline
with beard & CelebA     & 4.92\%  &  31,945                 & 4.41\%  & 164,409                  & 0.50\% \\
vs.        & FFHQ-Aging\footnotemark & NA      & NA                      & NA      & NA                       & NA \\
beardless  & synthetic  & 4.53\%  &   1,499                 & 3.53\%  &   4,021                  & 1.00\% \\
\hline
white skin & CelebA     & 4.24\%  & 127,647                 & 4.37\%  &  10,745                  & -0.12\% \\
vs.        & FFHQ-Aging & 1.68\%  &  26,984                 & 1.74\%  &   2,057                  & -0.07\% \\
black skin & synthetic  & 3.65\%  &   2,281                 & 4.45\%  &   1,184                  & -0.79\% \\
\hline
\bottomrule
\end{tabular}
\end{table}
\footnotetext{FFHQ-Aging does not have "beard" attribute annotation hence the corresponding row is empty.}

\section{Results And Conclusion}

Figure \ref{figure:face_images} shows three examples of faces images from each dataset. Table \ref{table:results_table} details the results of our experiments. Our results show that despite the clear domain gap between real and synthetic images, all the datasets show the same bias trend. This suggests either inherent difficulty in the less performing attribute or a bias embedded in the model itself due to training on biased data.

We conclude that synthetic face images generated by our platform can be used to uncover real weaknesses in an existing model trained for facial landmarks detection. As future research direction, we plan to conduct our experimental flow using other trained models as well as explore additional biases. Furthermore, we plan to show how adding carefully controlled synthetic data to the \textit{training} process can remedy these biases.

{
\small
\bibliographystyle{unsrt}
\bibliography{references}
}

\end{document}